\documentclass{article}
\usepackage{spconf,amsmath,graphicx}
\usepackage{booktabs}
\usepackage{amsmath}
\usepackage{tabularx}
\usepackage{multirow}
\usepackage{url}
\usepackage{adjustbox}
\usepackage[table]{xcolor}
\usepackage[bottom]{footmisc}
\usepackage{hyperref}
\usepackage{makecell}
\usepackage{caption}
\usepackage{stfloats}

\usepackage{cleveref}
\usepackage{array}
\newcolumntype{R}[1]{>{\raggedleft\arraybackslash}p{#1}}

\title{Swivuriso: The South African Next Voices Multilingual Speech Dataset}

\name{
\begin{tabular}{@{}c@{}}
Vukosi Marivate\textsuperscript{1,2,*} \quad Kayode Olaleye\textsuperscript{1,*} \quad Sitwala Mundia\textsuperscript{1} \quad Andinda Bakainga\textsuperscript{1} \\
Unarine Netshifhefhe\textsuperscript{1} \quad Mahmooda Milanzie\textsuperscript{1} \quad Tsholofelo Hope Mogale\textsuperscript{1} \quad Thapelo Sindane\textsuperscript{1}\\
Zainab Abdulrasaq\textsuperscript{1} \quad Kesego Mokgosi\textsuperscript{6} \quad Chijioke Okorie\textsuperscript{1,2} \quad Nia Zion Van Wyk\textsuperscript{1} \\
Graham Morrissey\textsuperscript{3} \quad Dale Dunbar\textsuperscript{3} \quad Francois Smit\textsuperscript{3} \quad Tsosheletso Chidi\textsuperscript{1} \\
Rooweither Mabuya\textsuperscript{4} \quad Andiswa Bukula\textsuperscript{5} \quad Respect Mlambo\textsuperscript{4} \quad Tebogo Macucwa\textsuperscript{1} \\
Idris Abdulmumin\textsuperscript{1} \quad Seani Rananga\textsuperscript{1}
\end{tabular}
}

\address{
\textsuperscript{1}University of Pretoria, South Africa \quad
\textsuperscript{2}AfriDSAI, University of Pretoria, South Africa \quad
\textsuperscript{3}Way With Words, South Africa \\
\textsuperscript{4}SADiLaR, South Africa \quad
\textsuperscript{5}Pennsylvania State University, USA \quad
\textsuperscript{6}Technological University Dublin, Ireland \\
\textsuperscript{*}\textit{Equal First Authors} \quad
\small{\textbf{Correspondence:} \href{mailto:vukosi.marivate@cs.up.ac.za}{vukosi.marivate@cs.up.ac.za}}
}

\begin{document}
\maketitle

\begin{abstract}
This paper introduces Swivuriso, a 3000-hour multilingual speech dataset developed as part of the African Next Voices project, to support the development and benchmarking of automatic speech recognition (ASR) technologies in seven South African languages. Covering agriculture, healthcare, and general domain topics, Swivuriso addresses significant gaps in existing ASR datasets. We describe the design principles, ethical considerations, and data collection procedures that guided the dataset creation. We present baseline results of training/finetuning ASR models with this data and compare to other ASR datasets for the languages concerned.
\end{abstract}


\section{Introduction}
\label{sec:intro}

Automatic speech recognition (ASR) has advanced significantly, mainly due to the emergence of deep learning methods and the increasing availability of extensive multilingual datasets~\cite{Dimah2024,KHEDDAR2024102422, Tjandra2023,Zeng2022,pmlr-v202-radford23a}. However, many African languages continue to be underrepresented in these advancements~\cite{Nakatumba-Nabende2024,elamin2023multilingual}. Limited availability of quality speech datasets for these languages restricts the creation and deployment of accurate and inclusive ASR systems~\cite{Etienne2014,BADENHORST2022107860}.

Current ASR data sets often do not reflect the rich linguistic and cultural diversity of the African continent~\cite{dossou2023advancing,chang2024self}. They are generally limited in scope~\cite{mak2024}, focusing predominantly on general-purpose speech. Moreover, these datasets are typically composed of scripted recordings, which limits the ability of ASR systems to handle spontaneous speech and multilingual interactions that are common in many African communities~\cite{BiswasYWWN20}.

\begin{table*}[t]
\centering
\small
\caption{Overview of selected South African ASR datasets. Swivuriso expands coverage in domain, scale, and speaker diversity.}
\begin{adjustbox}{width=\textwidth}
\begin{tabular}{@{}llllllll@{}}
\toprule
Dataset & \# Languages & Speech Type & Domain & License & Hours & \# Speakers & \# Clips \\
\midrule
NCHLT Speech~\cite{Etienne2014,BADENHORST2022107860}  & 10 & scripted & general & CC BY 3.0 & 560 & 2010 & 440738 \\
Lwazi ASR~\cite{gumede-plauche-2009-initial} & 10 & scripted & general, telephonic & CC BY 3.0 & 80 & 2012 & 53000+ \\
Common Voice 21.0~\cite{ardila-etal-2020-common} & 6 & scripted & general & CC-0 & 9 & 28 & 4404 \\
FLEURS~\cite{conneau2023fleurs} & 4 & scripted & general & CC BY & 48 & 13656 & 13656 \\
Vuk'uzenzele (ViXSD)~\cite{rajab2025esethu} & 1 & scripted & general & Esethu & 9 & 8 & 395 \\
\rowcolor{gray!20}
Swivuriso & 7 & scripted, unscripted & agri, general, health & CC BY 4.0 & 3000 & 2353 & 1M+ \\
\bottomrule
\end{tabular}
\end{adjustbox}
\label{tab:asr-dataset-summary}
\end{table*}

In response to these challenges, we introduce \emph{Swivuriso}%
\footnote{\emph{Swivuriso (tso)} means proverbs/sayings. The name reflects the dataset's role in unlocking knowledge in South African languages. See: \url{https://dsfsi.co.za/za-african-next-voices/}},
a comprehensive 3000-hour multilingual speech corpus developed as part of the African Next Voices (ANV) project. The dataset covers seven South African languages and focuses on three key domains: agriculture, healthcare, and general-purpose speech. It includes both scripted and unscripted speech, with prompts carefully designed to reflect culturally grounded and contextually relevant use. Developed through participatory methods and guided by strong ethical principles, Swivuriso aims to improve the representation of South African languages in ASR systems, promote community engagement, and uphold responsible data stewardship.

This paper contributes in four ways. We introduce Swivuriso, one of the largest publicly available multilingual speech datasets for South African languages; we provide broad coverage of underrepresented socioeconomic domains; we develop the dataset using community-centred, ethically grounded data-collection practices; and we present initial ASR benchmarking results that demonstrate the dataset's utility and indicate directions for future work. Together, these contributions establish Swivuriso as a significant resource for building inclusive and context-aware speech technologies in South Africa.

\section{Background and Related Work}
\label{sec:background}

\subsection{Existing South African Speech Datasets}
Only a handful of South African speech datasets have been publicly released to date, each differing in scale, domain focus, and how contributors were engaged---for instance, through open crowdsourcing or through organized community recruitment. Table~\ref{tab:asr-dataset-summary} provides an overview of publicly available South African speech datasets, including Swivuriso, and outlines their language coverage, data types, and domains. Among these, the NCHLT Speech Corpus and Lwazi ASR are the most extensive, covering ten South African languages. They primarily consist of scripted, read-aloud speech collected in controlled environments. While they offer broad linguistic coverage, they do not include spontaneous or domain-specific speech samples.

More recent datasets include Mozilla's Common Voice, which provides crowd-sourced, scripted speech in six South African languages, distributed under a permissive CC-0 license to encourage broad reuse. FLEURS includes approximately 12 hours of data per language in four South African languages, designed primarily for benchmarking and evaluation. Vuk'uzenzele (ViXSD)~\cite{rajab2025esethu} is a community-led ASR dataset, released under the Esethu licensing framework, emphasizing community ownership and ethical commercialization. While smaller in scale, it highlights local participation in resource creation. AfriSpeech-200~\cite{Afrispeech-200} includes English speech with South African accents and healthcare domain coverage, useful for accent-aware ASR modeling.

Across these resources, there is still limited availability of spontaneous speech and focused domain coverage. This highlights the need for more diverse, community-grounded datasets like Swivuriso that incorporate both scripted and unscripted speech, broader domain representation, and deeper speaker diversity.

\subsection{Ethical and Participatory Data Collection Practices}
Collecting speech data in underrepresented language communities demands sensitivity to consent, privacy, cultural norms, and long-term community impact~\cite{holton2021people,armentano-oller-etal-2024-becoming}. Prior work in projects such as Common Voice and ALFFA~\cite{besacier2015speech} has demonstrated the importance of embedding ethical and participatory practices throughout the data lifecycle. A recent case study on the Catalan edition of Common Voice highlights the need for clear consent procedures, privacy safeguards, and inclusive participation~\cite{armentano-oller-etal-2024-becoming}. Similarly, the ALFFA project emphasised community involvement by working with local experts and field linguists during data collection~\cite{besacier2015speech}.

Swivuriso builds on these previous efforts by emphasizing ethical practices and inclusive representation. Local coordinators played a central role in managing recordings, supporting participants, and ensuring community diversity. The data creation process followed institutionally approved ethical protocols, including anonymization of metadata, removal of personal identifiers, and safeguards to prevent misuse---such as voice cloning or biometric profiling.

\subsection{Licensing and Accessibility}
Licensing and accessibility are essential for ensuring the broad usability of datasets, particularly for low-resource languages. Open licenses---such as Creative Commons---have been widely recognized for enabling broad reuse, especially where data collection is limited or financially inaccessible~\cite{Pratap2020MLSAL}. Projects like Common Voice illustrate how permissive licensing can support equitable access and accelerate language technology development for underrepresented languages~\cite{ardila-etal-2020-common}.

In contrast, access-restricted datasets, while valuable, remain difficult to obtain for many researchers---particularly those based in low- and middle-income countries (LMICs). Yet, concerns around dataset accessibility are only part of a broader conversation. Increasingly, the focus is shifting toward ethical and equitable data governance, especially in contexts where language data is sourced from communities that have historically had limited control over how it is used. This has prompted interest in alternative licensing models that prioritize responsible stewardship, local agency, and benefit sharing~\cite{OkorieOminoLicensing,rajab2025esethu}.

Swivuriso draws on these principles, aiming to balance open access with ethical accountability. It adopts an open licensing approach while embedding guidance on responsible dataset usage.

\section{Swivuriso}
\label{sec:swivuriso}

\subsection{Dataset Scope and Languages}

\begin{figure}[!t]
    \centering
    \includegraphics[width=\linewidth]{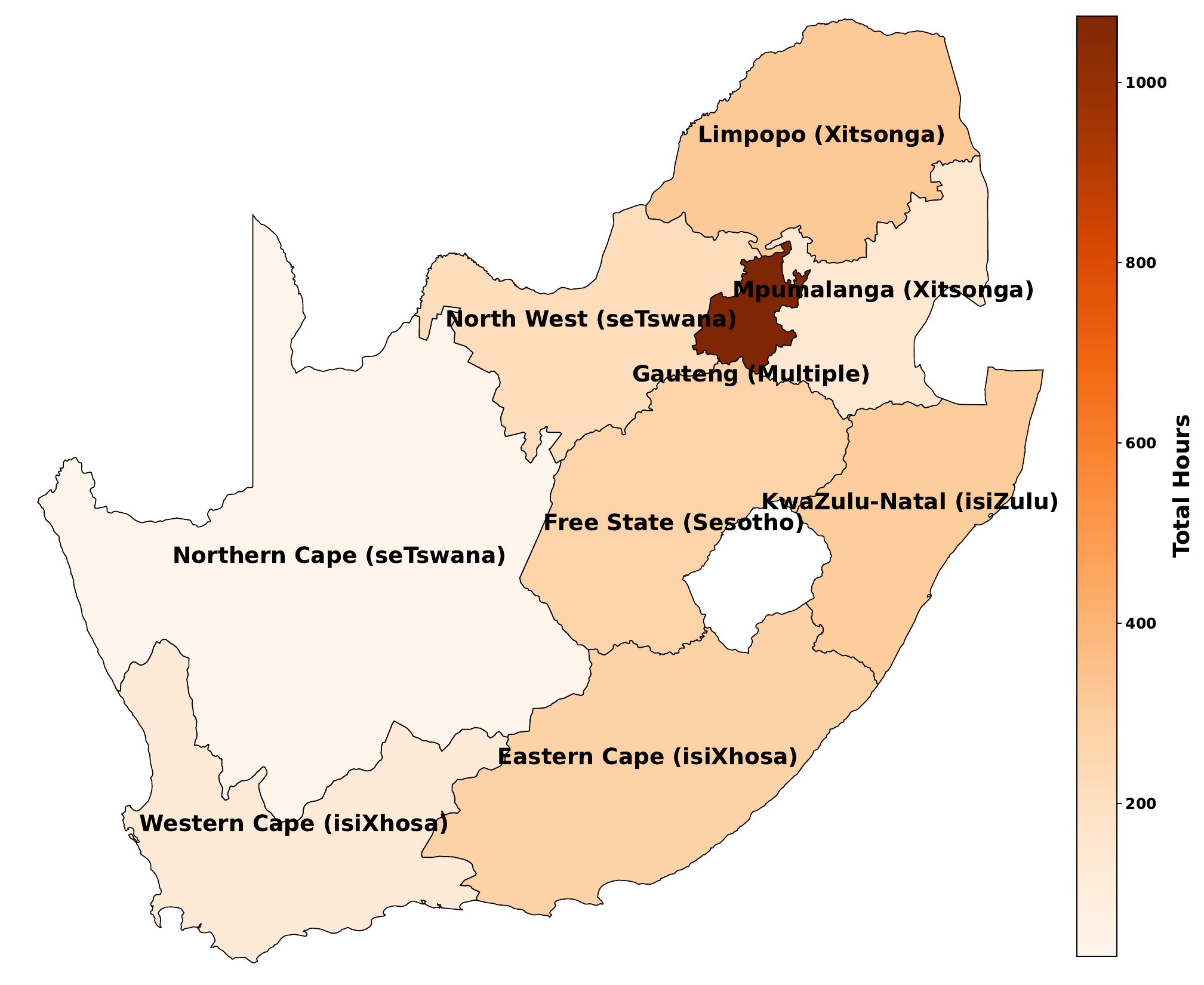}
    \caption{Geographic distribution of Swivuriso across South African provinces. Each province is shaded according to the total hours of recorded speech. The labels display the province name, with the dominant language indicated in brackets.}
    \label{fig:swivuriso_province_map}
\end{figure}

Swivuriso comprises approximately 3000 hours of speech collected from speakers of seven South African languages: isiZulu (ZUL), isiXhosa (XHO), Sesotho (SOT), Setswana (TSN), Xitsonga (TSO), Tshivenda (VEN), and isiNdebele (NBL). These languages were selected to reflect South Africa's rich linguistic diversity, spanning multiple regions and sociolinguistic contexts (see Figure~\ref{fig:swivuriso_province_map}).

\begin{table*}[t]
\centering
\setlength{\tabcolsep}{4pt}
\small
\caption{Overview of Swivuriso statistics per language showing total clips, number of speakers, hours of data, average durations, and clip-based demographic breakdowns (gender and age group values represent counts of clips).}
\begin{tabularx}{\textwidth}{@{} >{\raggedright\arraybackslash}X r r r r r r | *{5}{r} @{}}
\toprule
 &  &  &  &  &  &  & \multicolumn{5}{c}{\textbf{Age Group}} \\
\cmidrule(l){8-12}
\textbf{Language} & \textbf{\# Clips} & \textbf{\# Speakers} & \textbf{Hours} & \textbf{Avg Dur (s)} & \textbf{\# Male} & \textbf{\# Female} & \textbf{18--29} & \textbf{30--39} & \textbf{40--49} & \textbf{50--59} & \textbf{60+} \\
\midrule
ZUL  & 59,115 & 482 & 502.85 & 30.62 & 25,681 & 33,434 & 33,167 & 19,020 & 5,185 & 1,028 & 715 \\
XHO  & 73,665 & 485 & 504.28 & 24.64 & 29,094 & 44,571 & 38,476 & 24,716 & 8,831 & 1,211 & 431 \\
SOT  & 78,113 & 480 & 503.58 & 23.21 & 35,072 & 43,041 & 49,740 & 20,020 & 6,177 & 2,176 & 0 \\
TSN  & 99,527 & 487 & 502.18 & 18.16 & 29,424 & 70,103 & 68,025 & 26,185 & 4,828 & 489 & 0 \\
TSO  & 79,107 & 198 & 500.15 & 22.76 & 36,321 & 42,786 & 60,004 & 14,390 & 4,056 & 304 & 353 \\
VEN  & 42,402 & 104 & 250.89 & 21.30 & 18,065 & 24,337 & 33,761 & 6,697 & 1,268 & 380 & 296 \\
NBL  & 51,262 & 104 & 251.86 & 17.69 & 10,989 & 40,273 & 36,415 & 12,874 & 1,973 & 0 & 0 \\
\bottomrule
\end{tabularx}
\label{tab:swivuriso_stats}
\end{table*}

To support robust ASR development, the dataset includes both scripted and unscripted speech. Scripted prompts were curated to explicitly cover three domains---healthcare, agriculture, and general-purpose communication---and were adapted from publicly available, domain-relevant materials (see Section~\ref{ssec:scripted-prompts} for details). Unscripted prompts span a wider range of topics, including \textit{healthcare}, \textit{agriculture}, \textit{telecommunication}, \textit{finance}, and \textit{transport}. These were designed to elicit culturally grounded, spontaneous responses that capture naturalistic speech patterns.

Each participant contributed approximately one hour of audio, recorded across multiple sessions and under varied acoustic conditions. Associated speaker metadata includes \textit{age range} and \textit{gender}. Table~\ref{tab:swivuriso_stats} summarizes the dataset statistics per language, including total clips, number of speakers, hours recorded, average duration, and speaker demographics.

\subsection{Dataset Creation Process}

\begin{figure*}[t]
    \centering
    \includegraphics[width=\linewidth]{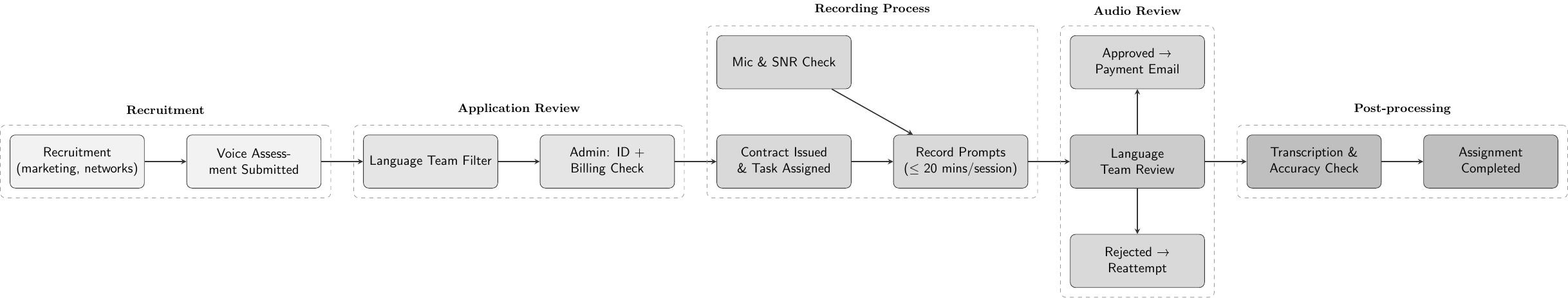}
    \caption{Overview of Swivuriso creation workflow.}
    \label{fig:data_collection_workflow}
\end{figure*}

The Swivuriso dataset was created through a structured, multi-stage workflow designed to ensure data quality and representative linguistic coverage (see Figure~\ref{fig:data_collection_workflow}). This process yielded over 3000 hours of validated scripted and unscripted speech across seven South African languages, covering key domains. The data collection pipeline was strictly phased:

\textbf{1. Recruitment and Screening:} Participants were sourced via targeted digital campaigns, community outreach, and university partnerships. Candidates completed a short voice-based screening to assess language proficiency and compliance with recording standards. Applications were reviewed by language-specific teams. To reduce fraudulent submissions, ID and facial verification were mandated.

\textbf{2. Onboarding and Setup:} Approved participants signed formal consent contracts and accessed TalkTag, a custom recording platform built for both scripted and unscripted prompts. Data collection primarily occurred in participants' home environments, contingent upon passing initial environmental noise and equipment checks.

\textbf{3. Recording Protocol:} All audio was captured in 48 kHz, 16-bit PCM WAV format to ensure high fidelity. Each session began with automated microphone and signal-to-noise ratio (SNR) checks. To manage vocal fatigue and preserve audio quality, individual sessions were strictly limited to 20 minutes, though contributors were permitted multiple sessions until a target duration of approximately one hour was reached.

\textbf{4. Quality Assurance and Transcription:} Submitted recordings were manually reviewed by language teams for technical clarity and adherence to task instructions. Approved audio was subsequently transcribed directly within the TalkTag platform by trained annotators. Only recording-transcript pairs that passed this final review were included in the dataset.

\textbf{5. Privacy, Ethics, and Remuneration:} Contributors provided informed consent and were instructed to avoid sharing personal content. Each utterance was linked to anonymized speaker metadata via a unique UUID, with all personally identifiable information (PII) encrypted and stored separately in line with ethical protocols. Remuneration was stratified by language to reflect specific engagement needs~\cite{de2017speech}.

\subsection{Prompt Design and Cultural Alignment}

\subsubsection{Unscripted Prompts}
\label{ssec:unscripted-prompts}
The unscripted prompts were created by dedicated language teams to ensure cultural relevance and naturalness. Each language team consisted of a language lead and language assistants (usually three members per team), the work within a given language was peer reviewed across the team before the language lead signed off on the prompts to be used within that project. Domain paths were set up via Microsoft Forms, whereby language teams would select a domain of knowledge to get started, narrowing down their focus by then selecting a topic within that domain of knowledge, which would then present further subtopics related to the topic at hand.

Language teams were trained to design prompts that sought to elicit a range of responses. A simple example would be: \textit{What is your favourite fruit, and why?} rather than \textit{What is your favourite fruit?} The language teams were trained to understand the importance of prompt design in gathering responses that were not only declarative, but also interrogative, exclamative, imperative, and so on. The teams were also urged to adjust topics and subtopics within a given domain to be more culturally relevant and, where necessary, sensitive. The goal was to capture the national lexicon of a language as far as possible, therefore, topics and subdomains were highly relevant to South African life and tailored towards the nuances that are specific to that language.

All prompts were composed directly in the first language (L1) of the respective language teams, although the prompt creation interface was presented in English. This ensured that no translation occurred in the process of generating unscripted prompts. Once a language team member had navigated to the end of a specific domain path, they would then formulate the prompt in L1. Before commencing prompt creation, all language team members completed a brief sample project on the system to fully understand the task. Subsequently, each language team generated a minimum of 2,000 unscripted prompts upon comprehending the project requirements and objectives.

\begin{table*}[t]
\centering
\small
\caption{Number of clips and unique speakers across train, dev, and devtest splits for each language.}
\begin{tabular}{l r r r r r r}
\toprule
\textbf{Language} & \multicolumn{2}{c}{\textbf{Train}} & \multicolumn{2}{c}{\textbf{Dev}} & \multicolumn{2}{c}{\textbf{Dev Test}} \\
\cmidrule(lr){2-3} \cmidrule(lr){4-5} \cmidrule(lr){6-7}
 & \textbf{Clips} & \textbf{Speakers} & \textbf{Clips} & \textbf{Speakers} & \textbf{Clips} & \textbf{Speakers} \\
\midrule
ZUL & 50,386 & 409 & 3,068 & 24 & 3,025 & 25 \\
XHO & 62,786 & 407 & 3,768 & 24 & 3,639 & 24 \\
SOT & 66,389 & 407 & 3,930 & 24 & 3,719 & 24 \\
TSN & 84,390 & 413 & 4,949 & 24 & 5,269 & 25 \\
TSO & 67,119 & 168 & 3,252 & 10 & 4,394 & 10 \\
VEN & 36,344 & 88  & 2,360 & 5  & 1,655 & 5  \\
NBL & 43,703 & 88  & 2,527 & 5  & 2,690 & 5  \\
\bottomrule
\end{tabular}
\label{tbl:clip_speaker_split_distribution}
\end{table*}

\begin{figure}
    \centering
    \includegraphics[width=\linewidth]{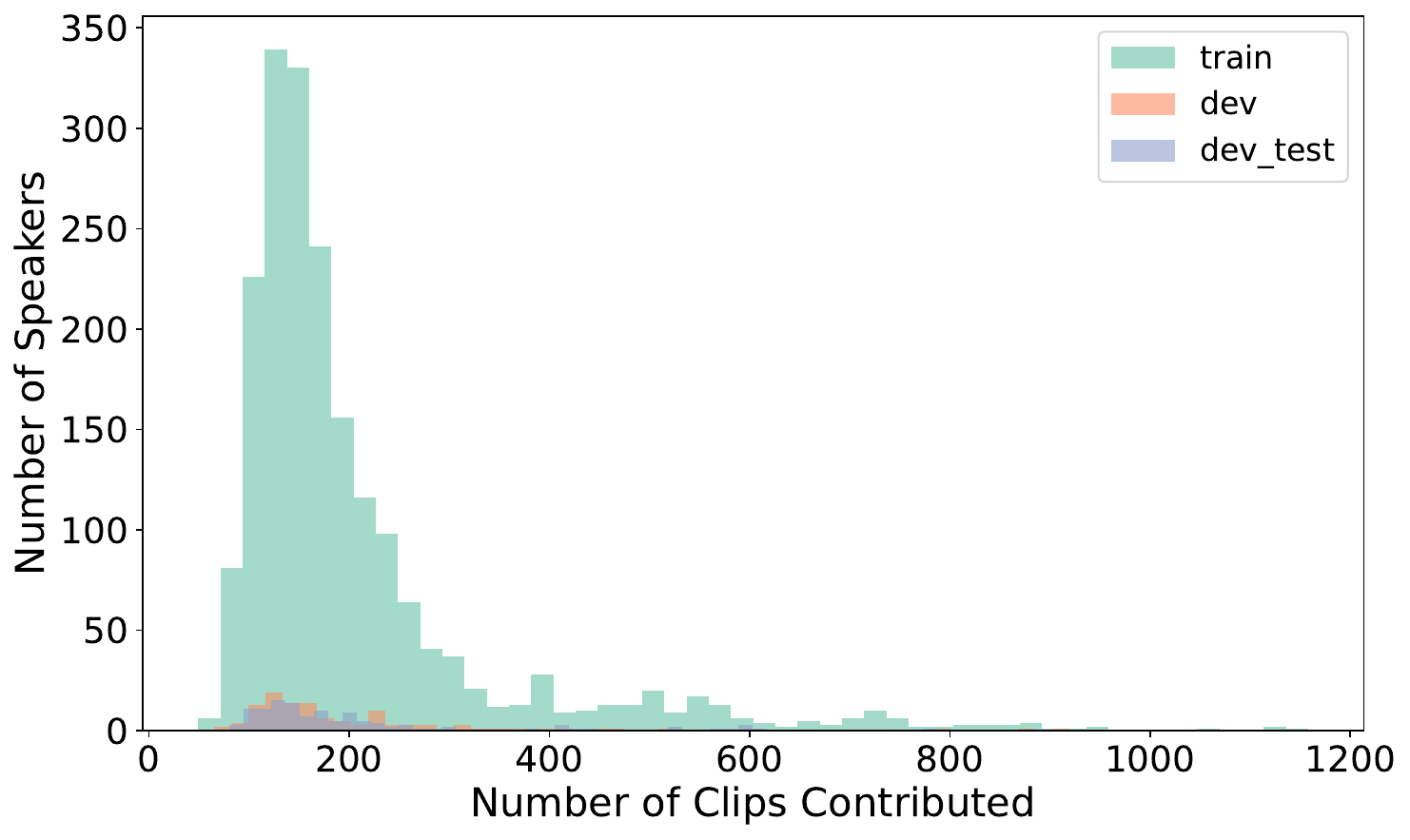}
    \caption{Histogram showing how many clips each speaker contributed in the train, dev, and devtest splits.}
    \label{fig:clip_per_speaker_distribution}
\end{figure}

\begin{figure}[!b]
    \centering
    \includegraphics[width=\linewidth]{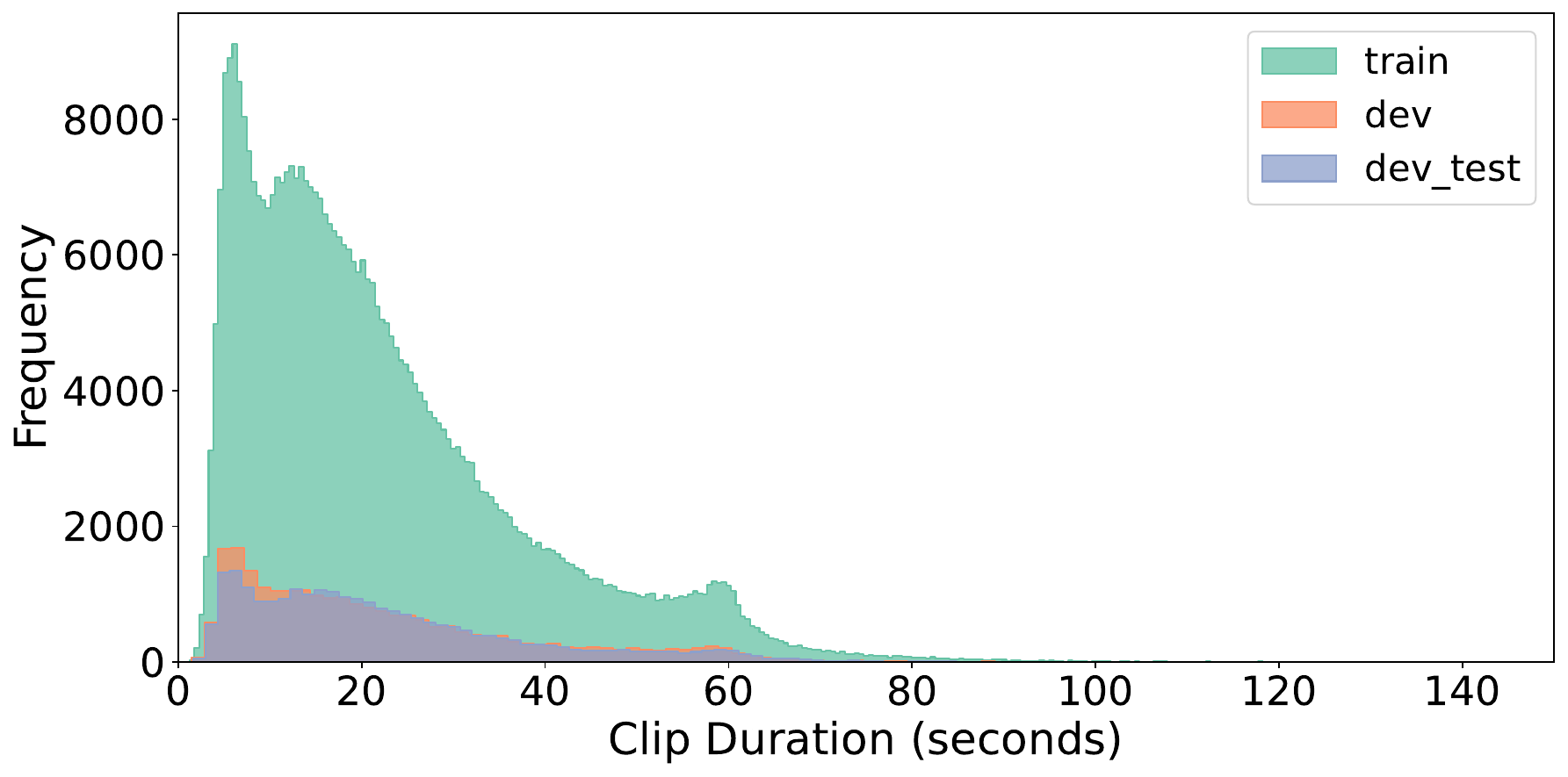}
    \caption{Distribution of clip durations across train, dev, and devtest splits. All splits display a highly similar distribution pattern, with the majority of clips concentrated between 10 and 30 seconds.}
    \label{fig:clips_duration_distribution}
\end{figure}

\begin{figure}
    \centering
    \includegraphics[width=\linewidth]{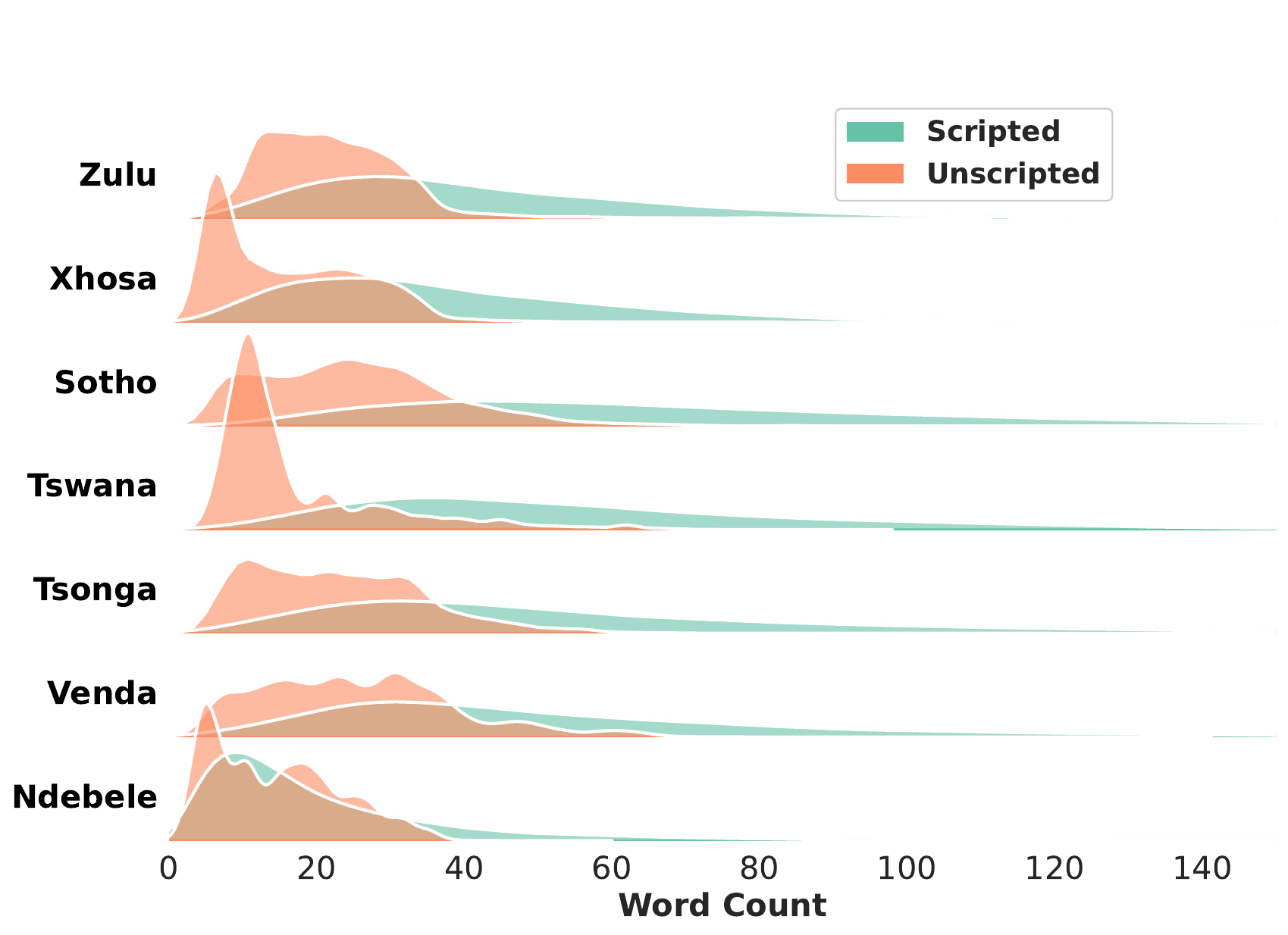}
    \caption{Distribution of word count per clip by speech type (scripted vs.\ unscripted) across languages. Each subplot displays histograms and KDE curves overlaid for both types, with vertical dashed lines marking median values.}
    \label{fig:word_count_by_type_per_language}
\end{figure}

\subsubsection{Scripted Prompts}
\label{ssec:scripted-prompts}
The scripted prompts were sourced from culturally relevant and domain-specific materials, ensuring diversity in areas such as agriculture and health. These sources included:
\begin{itemize}
	\item Pula/Imvula Magazine~\cite{grainsa_pula_imvula}, a monthly publication of Grain SA\footnote{\url{https://www.grainsa.co.za}} providing agricultural information in five South African languages (English, Sesotho, Setswana, isiZulu and isiXhosa). This data was made available through South African Centre for Digital Language Resources (SADiLaR)\footnote{\url{https://sadilar.org/}}.
	\item Vukuzenzele~\cite{lastrucci2023preparingvukuzenzelezagovmultilingualsouth}, a government publication that offers information on public services.
	\item Wikipedia (South African Languages), providing broad linguistic and thematic coverage.
	\item African Wordnet~\cite{sonja_marissa_wordnet}, a vital lexical database to ensure vocabulary accuracy and support the development of various language technologies in South African languages.
	\item Translated goatkeepers manuals (ARC animal health)~\cite{vatta2007Goatkeepers}, specialized content on animal health and agriculture.
\end{itemize}

The initial cleaning and preparation of the data for prompts involved a structured pipeline designed to ensure high-quality text suitable for training ASR models. Each stage refined the data, removing irrelevant content, categorizing text into themes, and identifying or flagging Personally Identifiable Information (PII).

Upon ingestion, initial quality control measures, including preliminary reviews for structural issues, were applied to these raw texts. The text was then filtered using language identification to retain only the target South African languages. While no strict confidence thresholds were applied, instances where the top-ranked language did not match the expected category were excluded. Confidence scores were retained as metadata.

To standardize the content and facilitate thematic analysis, two steps were performed: First, the entire text was translated into English using NLLB-200~\cite{nllbteam2022languageleftbehindscaling}. Second, translated text files were categorized into themes (e.g., agriculture, healthcare, general) using OpenAI's GPT-4~\cite{openai2023gpt4}.

Raw text often contains unnecessary information that was removed to standardize the dataset and improve the quality of the text. This involved the removal of metadata and irrelevant content, such as stripping HTML tags, URLs, excessive whitespace, and other non-textual artifacts. Custom cleaning rules were also developed and applied per source to address consistent issues found in specific datasets.

The text was segmented into meaningful sentences (prompts) using BlingFire~\cite{microsoft2021blingfire}. This ensured accurate segmentation by extracting discrete sentences, enforcing proper punctuation at each sentence end, and merging very short sentences to avoid fragmentation.

Finally, Named Entity Recognition (NER) and Personal Information Anonymization were applied. Specifically, Personally Identifiable Information (PII) was flagged or removed using Microsoft Presidio~\cite{microsoftPresidio}, which identifies various PII types.

\subsection{Dataset Statistics}

\begin{table*}[ht!]
\centering
\caption{Distribution of total hours recorded across different content domains, grouped by train, dev, and devtest splits.}
\begin{tabular}{l||l R{1.2cm}R{1.2cm}R{1.2cm} || l R{1.2cm}R{1.2cm}R{1.2cm}}
\toprule
&  & \multicolumn{3}{c||}{\textbf{Number of Hours}} & & \multicolumn{3}{c}{\textbf{Number of Hours}} \\ \cmidrule(lr){3-5}\cmidrule(lr){7-9}
\textbf{Domain} & \textbf{Lang.} & \textbf{Train} & \textbf{Dev} & \textbf{DevTest} & \textbf{Lang.} & \textbf{Train} & \textbf{Dev} & \textbf{DevTest} \\
\midrule
Agriculture & ZUL & 76.24 & 4.42 & 4.52 & XHO & 76.58 & 4.33 & 4.60 \\
Culture and Society & & 36.40 & 2.21 & 2.16 &  & 36.09 & 2.10 & 2.14 \\
Finance & & 36.54 & 2.17 & 2.19 &  & 36.72 & 2.23 & 2.23 \\
General & & 92.27 & 5.34 & 5.64 &  & 92.74 & 5.47 & 5.45 \\
Health & & 76.22 & 4.28 & 4.66 &  & 76.21 & 4.41 & 4.56 \\
Sports and Hobbies & & 36.53 & 2.15 & 2.08 &  & 36.18 & 2.19 & 2.16 \\
Telecommunication & & 36.70 & 2.18 & 2.24 &  & 36.53 & 2.24 & 2.26 \\
Transport & & 36.49 & 2.14 & 2.18 &  & 36.35 & 2.21 & 2.20 \\
\midrule
Agriculture & SOT & 77.65 & 4.56 & 4.37 & TSN & 75.19 & 4.41 & 4.64 \\
Culture and Society & & 35.66 & 2.10 & 2.07 &  & 35.96 & 2.12 & 2.16 \\
Finance & & 36.37 & 2.10 & 2.26 &  & 36.41 & 2.14 & 2.20 \\
General & & 92.20 & 5.40 & 5.44 &  & 93.29 & 5.44 & 5.59 \\
Health & & 75.96 & 4.49 & 4.41 &  & 75.64 & 4.43 & 4.62 \\
Sports and Hobbies & & 36.54 & 2.15 & 2.20 &  & 36.56 & 2.14 & 2.16 \\
Telecommunication & & 36.77 & 2.13 & 2.21 &  & 36.58 & 2.11 & 2.20 \\
Transport & & 36.44 & 2.08 & 2.14 &  & 36.18 & 2.15 & 2.21 \\
\midrule
Agriculture & TSO & 72.32 & 3.51 & 4.31 & NBL & 35.51 & 2.43 & 2.43 \\
Culture and Society & & 41.13 & 2.05 & 2.49 &  & 19.78 & 1.36 & 1.28 \\
Finance & & 41.69 & 2.07 & 2.51 &  & 20.39 & 1.39 & 1.36 \\
General & & 35.97 & 1.56 & 2.20 &  & 14.70 & 1.07 & 1.04 \\
General - Civic Life & & 16.68 & 0.71 & 0.98 &  & 5.51 & 0.38 & 0.38 \\
General - Knowledge & & 18.83 & 0.82 & 1.24 &  & 9.29 & 0.68 & 0.68 \\
Health & & 76.36 & 3.74 & 4.47 &  & 44.16 & 3.01 & 2.99 \\
Sports and Hobbies & & 41.69 & 2.06 & 2.51 &  & 20.23 & 1.36 & 1.36 \\
Telecommunication & & 41.55 & 2.02 & 2.48 &  & 20.18 & 1.37 & 1.34 \\
Transport & & 41.87 & 2.05 & 2.47 &  & 20.08 & 1.40 & 1.32 \\
\midrule
Agriculture & VEN & 40.95 & 2.40 & 2.26 & & & & \\
Culture and Society & & 20.65 & 1.22 & 1.10 & & & & \\
Finance & & 21.22 & 1.21 & 1.12 & & & & \\
General & & 13.61 & 0.81 & 0.75 & & & & \\
General - Civic Life & & 7.21 & 0.41 & 0.31 & & & & \\
General Topics & & 7.32 & 0.42 & 0.31 & & & & \\
Health & & 40.81 & 2.39 & 2.29 & & & & \\
Sports and Hobbies & & 20.93 & 1.20 & 1.12 & & & & \\
Telecommunication & & 20.90 & 1.21 & 1.10 & & & & \\
Transport & & 20.89 & 1.20 & 1.12 & & & & \\
\bottomrule
\end{tabular}
\label{tab:domain_hours_distribution}
\end{table*}

To support fair and consistent evaluation, Swivuriso is released with predefined \textit{train}, \textit{dev}, and \textit{devtest} splits, following an 85/5/5 split. The actual hours per split is presented in \Cref{tab:domain_hours_distribution}. An additional 5\% of the data is held out as a private \textit{test} set to support future shared tasks and public ASR leaderboard evaluation.

These splits are speaker-disjoint, ensuring no speaker appears in more than one subset. This ensures reliable benchmarking and reproducibility across studies, while reducing the risk of data leakage.

Figure~\ref{fig:clip_per_speaker_distribution} compares the number of clips and unique speakers across train, dev and test splits for each language. The distributions closely follow the expected split ratio, with the training split containing the majority of data. Figure~\ref{fig:clips_duration_distribution} shows the distribution of clip durations across splits. Most clips fall between 10 and 30 seconds, and the shapes of the distributions are nearly identical across splits. \Cref{tab:domain_hours_distribution} shows the total number of hours recorded per domain across train, dev, and test splits. All domains are represented in each split, with the training set holding the majority of the content, as expected. While \textit{general}, \textit{health}, and \textit{agriculture} dominate in terms of hours, other domains such as \textit{finance}, \textit{telecommunication}, and \textit{culture and society} also have meaningful representation. \Cref{fig:word_count_by_type_per_language} shows the distribution of word count per clip for scripted and unscripted recordings. In all languages, scripted clips tend to have much shorter word counts, with medians ranging between 17--26 words. Unscripted clips are consistently longer, with higher medians (33--75 words) and more variability. The histograms show that scripted speech distributions are narrow and sharply peaked, reflecting uniformity in read prompts, while unscripted speech exhibits longer tails, capturing the natural variation of spontaneous speech.

The dataset is publicly available through multiple mirrors to support long-term accessibility and reuse. The primary dataset repository, including documentation and data processing scripts, is hosted on GitHub\footnote{\url{https://github.com/dsfsi/za-african-next-voices}}. The full (uncompressed) dataset can also be accessed and explored on HuggingFace\footnote{\url{https://huggingface.co/datasets/dsfsi-anv/za-african-next-voices}}. A citable, compressed, archived version is preserved on Zenodo\footnote{\url{https://zenodo.org/records/17776289}}, ensuring persistence and versioning for research use. Additional project information, updates, and related resources are available on the DSFSI project website\footnote{\url{https://www.dsfsi.co.za/za-african-next-voices/}}.

\section{Automatic Speech Recognition Benchmarks}
\label{sec:asr}

We demonstrate the potential of the Swivuriso dataset by performing automatic speech recognition (ASR) in seven South African languages. A series of experiments were conducted to benchmark model performance and establish baselines on the Swivuriso dataset. Prior to fine-tuning, we resampled all audio recordings to 16 kHz and applied standard preprocessing to the transcripts, which included normalizing the text, and removing both punctuation and speech tags. For the experiments, we used the following models: whisper-large-v3-turbo (809M), Wav2vec-BERT 2.0 (600M) and MMS-1b-all (1B). All experiments were conducted on a compute node featuring 24 vCPUs and 170 GB of system memory, equipped with 2$\times$ NVIDIA A100 GPUs (40GB each).

\subsection{South African Cross-Corpus Performance}
First, a comparison was made between the performance of several models on public ASR datasets (Lwazi, NCHLT) and the Swivuriso dataset. To ensure a fair comparison, larger datasets were downsampled to align with the smallest dataset, Lwazi, resulting in an average of approximately six hours of audio per language. Both Wav2vec-BERT and MMS were then fine-tuned on these respective datasets.

\vspace{3mm}
\textbf{Training Configuration:} All training was conducted for 2000 steps with an effective batch size of 32 and learning rates of $5 \times 10^{-5}$ and $1 \times 10^{-3}$ for Wav2vec-BERT and MMS respectively.

\vspace{3mm}
\textbf{Results:} The Word Error Rates (WERs) for the selected models are presented in Table~\ref{tab:wer_results}. Although absolute WERs on the Swivuriso sample are slightly higher than those on the NCHLT sample, this discrepancy highlights the increased acoustic and linguistic complexity inherent in the Swivuriso recordings. This is empirically confirmed by the cross-corpus analysis below. These results underscore the necessity of developing contemporary, high-volume language resources that capture a broader distribution of speech patterns than legacy datasets.

\begin{table}[h]
\centering
\caption{Comparison of Word Error Rates (WER) across datasets.}
\begin{tabular}{lcc}
\toprule
\textbf{Dataset} & \textbf{\makecell{Wav2Vec2-BERT\\2.0 (600M)}} & \textbf{MMS (1B)} \\
\midrule
Lwazi & 0.45 & 0.35 \\
NCHLT Sample & 0.36 & 0.41 \\
Swivuriso Sample & 0.38 & 0.45 \\
\bottomrule
\end{tabular}
\label{tab:wer_results}
\end{table}

\vspace{3mm}
\noindent\textbf{Whisper Cross-Corpus Generalization:} To further investigate dataset generalizability, we conducted a second set of experiments using the whisper-large-v3-Turbo architecture. Two separate models were fine-tuned on the seven languages to assess how well a model trained on one corpus performs when evaluated on the other. One model was trained exclusively on the Swivuriso dataset, while the second was trained on the NCHLT dataset.

\vspace{3mm}
\noindent\textbf{Training Configuration (Whisper):} The models were fine-tuned using Whisper-large-v3-Turbo (809M). All training was conducted for 2000 steps with an effective batch size of 64 and a learning rate of $1 \times 10^{-5}$.

\vspace{3mm}
\noindent\textbf{Cross-Corpus Results:} The cross-evaluation results are detailed in Table~\ref{tab:whisper_cross_results}. While both models perform best on in-domain data, a significant performance asymmetry is observed in their generalization capabilities. The model trained on NCHLT shows a catastrophic increase in WER when evaluated on Swivuriso (+0.57 $\Delta$). Conversely, the Swivuriso-trained model demonstrates high robustness, with a minimal degradation of only 0.05 absolute WER when applied to the NCHLT test set. This small transfer gap suggests that Swivuriso acts as a linguistic superset, capturing a broader distribution of South African speech patterns that enables superior generalization compared to legacy datasets.

\begin{table}[h]
    \centering
    \small
    \caption{Cross-corpus evaluation. While NCHLT is ``easier'' in-domain, it fails to generalize to Swivuriso (+0.57 $\Delta$); conversely, Swivuriso demonstrates high robustness with minimal degradation when tested on NCHLT (+0.05 $\Delta$).}
    \begin{tabularx}{\columnwidth}{lXX}
        \toprule
        & \multicolumn{2}{c}{\textbf{Evaluation Set (WER)}} \\
        \cmidrule(lr){2-3}
        \textbf{Train Set} & \textbf{Swivuriso} & \textbf{NCHLT} \\
        \midrule
        Swivuriso & \textbf{0.24} & 0.29 \\
        NCHLT     & 0.71 & \textbf{0.14} \\
        \bottomrule
    \end{tabularx}
    \label{tab:whisper_cross_results}
\end{table}

\subsection{Whisper-large-v3-Turbo Monolingual Baselines}

\begin{figure*}[h]
\centering
\includegraphics[width=0.7\linewidth]{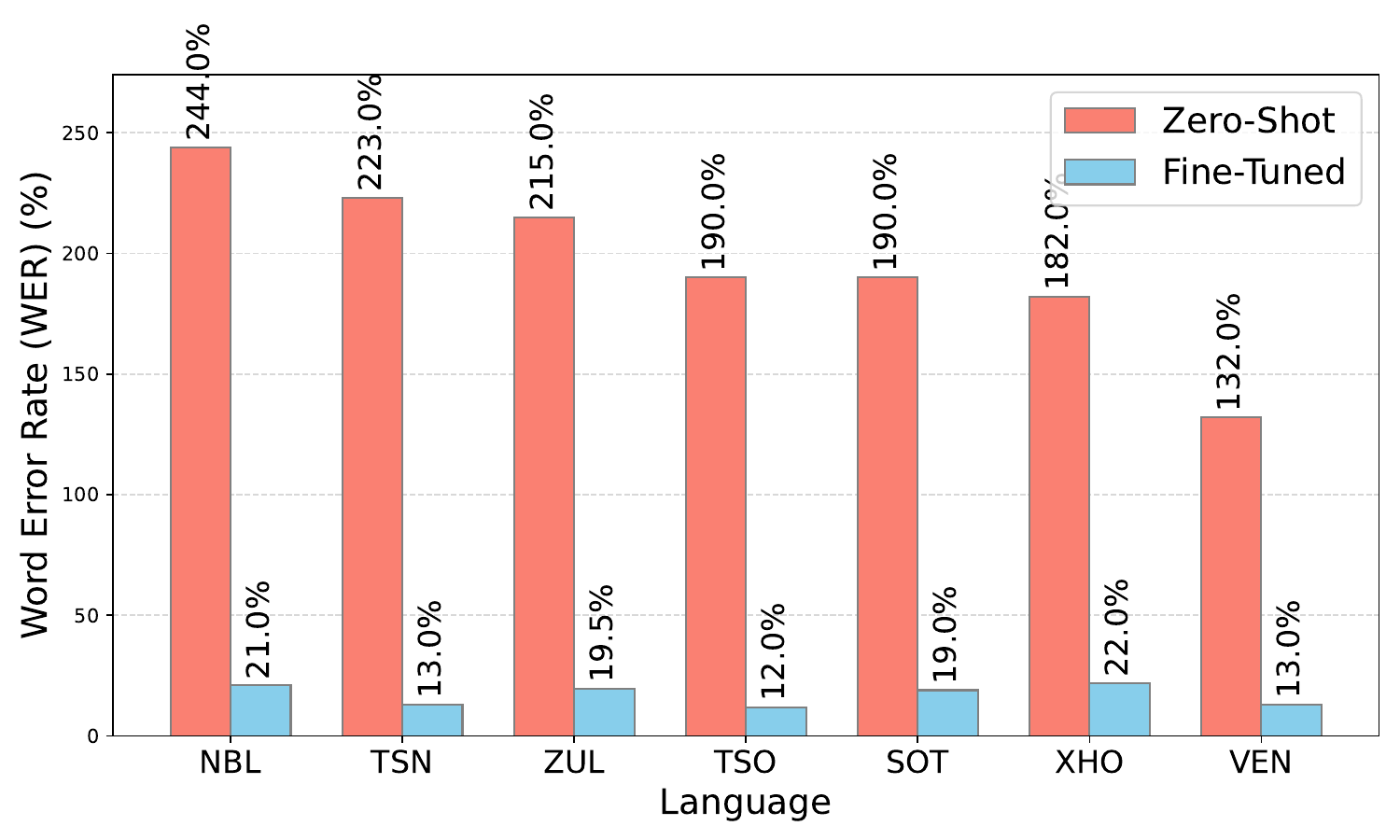}
\captionsetup{width=0.7\linewidth}
\caption{Comparison of Word Error Rates (WER) on the Swivuriso dataset. The Whisper-large-v3-turbo model was evaluated in Zero-Shot mode and after fine-tuning for 2000 steps.}
\label{fig:wer_comparison}
\end{figure*}

Baseline models were fine-tuned on the Swivuriso dataset using Whisper-large-v3-Turbo (809M), a pruned version of Whisper Large.

\vspace{3mm}
\textbf{Training Configuration:} All training was conducted for 2000 steps with an effective batch size of 64 and a learning rate of $1 \times 10^{-5}$.

\vspace{3mm}
\textbf{Results:} Both zero-shot and fine-tuned performance were evaluated. As shown in Figure~\ref{fig:wer_comparison}, fine-tuning resulted in a reduction in WER across all languages. Prior work has documented that Whisper exhibits hallucination behavior in low-resource language settings, which can produce insertion-heavy errors and word error rates above 100\%~\cite{Bara_ski_2025}. Consistent with this, zero-shot WER for SeTswana (TSN) was 223.0\%, which dropped to 13.0\% after fine-tuning. Similarly, XiTsonga (TSO) improved from 190.0\% to 12.0\%.

\vspace{3mm}
\noindent\textbf{Scripted vs.\ Unscripted Performance:} We further investigated the model's performance differences between scripted and unscripted speech styles. Table~\ref{table:scripted_unscripted} details the Word Error Rate (WER) for the whisper-large-v3-Turbo model fine-tuned with the same training configuration.

While unscripted speech generally yields a lower mean WER, this appears to be driven by a bimodal error distribution where simpler, natural utterances achieve near-perfect transcription. In contrast, the scripted subset exhibits a higher baseline error rate, largely due to quality control issues inherent in web-scraped data. Unlike the curated unscripted set, the scripted references suffer from non-standard orthography and severe language alignment issues.

Table~\ref{table:scripted_errors} highlights these failure modes. The first example demonstrates morphological errors, where the model misses the prefix in the agglutinative target language. The second example illustrates a critical ``Language Mismatch'' issue common in scraped datasets: the reference text is in English (likely the original article source), but the audio contains the speaker translating or code-switching into Zulu (e.g., saying ``\textit{ekhombisa}'' instead of ``\textit{which shows}''). Consequently, the model is penalized for correctly transcribing the speech because it deviates from the mismatched English ground truth.

\begin{table}[h]
\centering
\small
\caption{Performance of fine-tuned whisper-large-v3-Turbo on Scripted vs.\ Unscripted speech test sets.}
\label{table:scripted_unscripted}
\begin{tabular*}{1\linewidth}{l @{\extracolsep{\fill}} r r}
\toprule
\textbf{Language} & \multicolumn{2}{c}{\textbf{WER (\%)}} \\
\cmidrule(lr){2-3}
 & \textbf{Scripted} & \textbf{Unscripted} \\
\midrule
ZUL & 20.93 & 17.79 \\
SOT & 22.56 & 15.26 \\
VEN & 6.69 & 15.21 \\
MULTILINGUAL & 18.21 & 15.76 \\
\bottomrule
\end{tabular*}
\end{table}

\begin{table}[h]
\centering
\small
\renewcommand{\arraystretch}{1.2}
\caption{Analysis of error types in Scripted data. The dataset suffers from morphological mismatches (top) and language alignment artifacts from scraping (bottom).}
\label{table:scripted_errors}
\begin{tabularx}{\linewidth}{X X l}
\toprule
\textbf{Reference} & \textbf{Prediction} & \textbf{Error Type} \\
\midrule
ziningi izindaba zezinkampani ezikwazile\ldots & ziningi izindaba \textbf{zenkampani} ezikwazile\ldots & \textit{Morphological} \\
 & & (High WER / Low CER) \\
\midrule
\ldots reality show \textbf{which shows internet sensation} zodwa wabantu & \ldots reality show \textbf{ekhombisa umthokozelwi} internet uzodwa wabantu & \textit{Language Mismatch} \\
 & & (English Ref vs.\ Zulu Audio) \\
\bottomrule
\end{tabularx}
\end{table}

\subsection{Multilingual Baseline Experiments}

We establish baseline performance on the Swivuriso dataset by fine-tuning three speech recognition architectures: Whisper-large-v3-Turbo, Wav2Vec-BERT 2.0, and Massively Multilingual Speech (MMS).

\vspace{3mm}
\textbf{Training Configuration:}
All models were fine-tuned for 10{,}000 optimisation steps with an effective batch size of 32, achieved through gradient accumulation. We employed model-specific learning rates: $1 \times 10^{-5}$ for Whisper, $5 \times 10^{-5}$ for Wav2Vec-BERT, and $3 \times 10^{-4}$ for MMS. Training was conducted using mixed-precision (bfloat16). For MMS, we fine-tune adapter layers, whilst Whisper and Wav2Vec-BERT are fine-tuned end-to-end with gradient checkpointing enabled for memory efficiency.

\begin{table}[h]
\centering
\caption{Baseline multilingual ASR results on Swivuriso. WER is reported on the held-out test set at 1{,}000 and 10{,}000 training steps. Lower is better.}
\begin{tabular}{lcc}
\toprule
\textbf{Model} & \textbf{1K Steps} & \textbf{10K Steps} $\downarrow$ \\
\midrule
Whisper-large-v3-Turbo & 0.37 & \textbf{0.15} \\
Wav2Vec-BERT 2.0 & 0.44 & 0.17 \\
MMS-1B-all & 0.47 & 0.32 \\
\bottomrule
\end{tabular}
\label{tab:mbaseline_results}
\end{table}

\vspace{3mm}
\textbf{Results:}
Table~\ref{tab:mbaseline_results} summarises the Word Error Rate (WER) achieved by each model at early (1{,}000 steps) and final (10{,}000 steps) training checkpoints. Continued fine-tuning yields substantial improvements across all models, with Whisper-large-v3-Turbo achieving the largest absolute reduction of 0.22, followed by Wav2Vec-BERT at 0.27 and MMS at 0.15.

\subsection{Effect of Speech Hours in Training Data}

To understand the impact of data volume on the fine-tuning process, an investigation into the effect of speech hours in the training data was conducted. The whisper-v3-turbo model was trained for a single epoch using subsets of 50 hours, 150 hours, and 250 hours of training data.

\vspace{3mm}
\textbf{Training Configuration:} All training was 1 epoch with an effective batch size of 64 and a learning rate of $1 \times 10^{-4}$.

\vspace{3mm}
\textbf{Results:} Table~\ref{table:training_hours_experiment} details the Word Error Rate (WER) and Character Error Rate (CER) relative to recorded hours in the train set. Increasing the training data volume generally leads to improved performance (lower WER and CER). For instance, XHO (Xhosa) WER improved from 39.63\% at 50 hours to 24.15\% at 250 hours. Performance for some languages, such as VEN (Venda), appeared to plateau after 150 hours.

\begin{table*}[t]
\centering
\small
\captionsetup{width=0.7\linewidth}
\caption{WER and CER performance across recorded hours in training data for each language. Whisper-large-v3-turbo fine-tuned for 1 epoch.}
\begin{tabular*}{0.7\linewidth}{l @{\extracolsep{\fill}} r r r r r r}
\toprule
\textbf{Language} & \multicolumn{3}{c}{\textbf{WER (\%)}} & \multicolumn{3}{c}{\textbf{CER (\%)}} \\
\cmidrule(lr){2-4} \cmidrule(lr){5-7}
 & \textbf{50hrs} & \textbf{150hrs} & \textbf{250hrs} & \textbf{50hrs} & \textbf{150hrs} & \textbf{250hrs} \\
\midrule
ZUL & 31.70 & 23.00 & 22.79 & 7.55 & 5.10 & 5.08 \\
XHO & 39.63 & 26.08 & 24.15 & 12.27 & 6.97 & 5.73 \\
SOT & 24.47 & 19.79 & 17.05 & 11.39 & 9.45 & 8.03 \\
TSN & 20.09 & 14.93 & 13.26 & 8.37 & 6.49 & 6.63 \\
TSO & 24.60 & 17.44 & 12.65 & 9.92 & 7.99 & 4.97 \\
VEN & 17.64 & 13.89 & 14.00 & 4.23 & 3.31 & 3.36 \\
NBL & 30.37 & 22.23 & 21.21 & 9.72 & 6.14 & 5.93 \\
\bottomrule
\end{tabular*}
\label{table:training_hours_experiment}
\end{table*}

\subsection{Discussion}

The baseline experiments show that the Swivuriso dataset supports measurable improvements in ASR performance across multiple architectures and languages. Fine-tuning consistently reduced error rates, and results across models indicate that the dataset provides stable learning signals even at early training stages. The analysis of training hours further confirms that additional data contributes to performance gains for most languages. These findings position Swivuriso as a reliable resource for future work on South African ASR and for continued methodological development in multilingual speech modelling.

\section{Related Work}
\label{sec:related}

There has been substantial work done previously by many scholars in the art of collecting and designing world class datasets especially in the field of Automatic Speech Recognition where the requirements for precision and elimination of anomalies are very important~\cite{junczyk2024survey, Afrispeech-200, kahn2020libri, yeroyan2024enabling, virkkunen2023finnish, kolobov2021mediaspeech}. In the context of low resource languages globally the development has been slow particularly for African languages. We take note of the ones that has helped enlighten us in this journey. Abdulmumin et al.\ proposed a study~\cite{abdulmumin2024correcting} on the FLORES dataset which helped correct FLORES to better evaluate four African low resource languages. This work involved a meticulous review and correction of the dataset's dev and devtest splits for Hausa, Northern Sotho (Sepedi), Xitsonga, and isiZulu. In this study, native speakers carefully identified and fixed inconsistencies and inaccuracies that could impact the reliability of downstream NLP tasks, especially machine translation. The work further demonstrated that their corrections improved the linguistic accuracy and overall quality of the dataset, supported by both qualitative assessment and statistical analysis of the changes. This correction effort highlights the importance of native speaker involvement to ensure linguistic and cultural relevance in low-resource language datasets. We have heavily involved native speakers in our dataset to strengthen its character and overall improve the linguistic accuracy and the overall quality of the dataset as recommended by~\cite{abdulmumin2024correcting}. Another interesting study which helped guide our process in terms of sentiment analysis across a range of domains is the study proposed by~\cite{ekbal2022hindimd}. The study identified that many sentiment analysis benchmark datasets for high resource languages have been made available; however, for low-resource languages this is not the case especially for Indian languages like Hindi. The study addressed the challenges of sentiment analysis in Hindi particularly addressing socially relevant domains by introducing a balanced corpus annotated with the sentiment classes, viz.\ positive, negative and neutral.

\section{Dataset Availability and Licensing}
\label{sec:licensing}

Dataset licensing is fundamental for ensuring both broad usability and ethical access, especially for low-resource languages. Standard open licenses, such as Creative Commons (e.g., CC BY 4.0), have made large-scale reuse possible, with initiatives like Mozilla Common Voice demonstrating the value of permissive licensing for language technology development. However, access restrictions to resources for AI development continue to impede many researchers in low- and middle-income countries, pointing to broader concerns about equitable data governance and local community agency~\cite{OkorieOminoLicensing}.

For the Swivuriso project, we adopted the Creative Commons Attribution 4.0 International (CC BY 4.0) license to promote open access while embedding ethical accountability. Swivuriso is released on Zenodo and Hugging Face, with persistent DOIs and comprehensive documentation. Each release includes audio, aligned transcripts, and detailed metadata on demographic, linguistic, and accessibility factors. We also provide an ethical use statement, restricting uses such as voice cloning to protect speaker privacy.

The CC BY 4.0 license is highly permissive: users can share, adapt, and build upon the dataset for any purpose, including commercially, provided they attribute the creators and indicate any modifications. The license prohibits additional legal or technical restrictions, covers database rights, and includes a ``cure period'' for addressing license violations, all designed to encourage legal certainty and global scientific collaboration.

However, while open licenses are essential for maximising data utility, they often fail to address issues like local agency or protection of Indigenous knowledge. The uniform application of standard licenses can unintentionally undermine community ownership, especially for cultural materials or community-sourced data. This limitation is driving a shift toward more context-sensitive frameworks that prioritize stewardship, benefit-sharing, and community control---an approach that is gaining ground in African data governance~\cite{rajab2025esethu,OkorieOminoLicensing}.

Recent licensing models such as the Nwulite Obodo Open Data License (NOODL)~\cite{okorie2025s,okorie2025addressing} and the Esethu License~\cite{rajab2025esethu} illustrate this paradigm shift. NOODL, for example, differentiates between users in Africa or developing countries and those outside these regions, enabling tailored benefit-sharing and requiring adaptations to be shared within the region. Esethu reinvests licensing revenue directly into expanding resources and supporting local employment, maintaining both community agency and resource sustainability.

As experience with Indigenous and low-resource language data has shown, unrestricted ``openness'' can reinforce existing inequities if not balanced with robust ethical safeguards. The current evolution of open data practice acknowledges that responsible sharing requires more than a generic license; it demands attention to consent, privacy, fairness, and mechanisms for ongoing community benefit.

Our choice to use CC BY 4.0 for Swivuriso was rooted in a desire for openness, but it also underscores the shortcomings of standard licenses for datasets from African and low-resource language communities. Emerging frameworks like NOODL and Esethu highlight the field's move towards more ethical, equitable governance. For dataset publications, it is no longer sufficient to declare a standard open license; authors should also explain their approach to ethical, just, and community-empowering data stewardship.

\section{Limitations}
\label{sec:limitations}

While this work establishes baseline results, several experimental limitations should be noted. Due to computational constraints, fine-tuning was conducted on a best-effort basis for a fixed number of steps rather than until full convergence. These resource limitations also precluded an exhaustive hyperparameter grid search. Furthermore, we prioritized establishing a broad benchmark across all seven languages in the Swivuriso corpus to establish a representative South African speech baseline. Consequently, intensive long-form training for individual languages was precluded, leaving room for further optimization and experimentation in future research.

\section{Conclusions}
\label{sec:conclusion}

Swivuriso represents a significant step toward advancing automatic speech recognition and broader speech technologies for South African languages. With coverage of seven linguistically and geographically diverse languages, and the inclusion of both domain-specific and culturally grounded unscripted speech, the dataset can support a wide range of use cases---including health communication, agricultural advisory tools, and multilingual spoken language applications.

The dataset is designed to support not only benchmark-driven evaluation---such as improving word error rate and related performance metrics---but also the development of speech systems that are culturally informed, socially responsive, and better aligned with the linguistic realities of underrepresented communities.

By grounding the dataset in community-informed consent, open licensing, and accessible distribution, the South Africa NextVoices project offers a replicable model for ethically robust and practically useful dataset creation in low-resource settings.

Future extensions may involve expanding the dataset to include additional Southern African languages and enriching it with task-oriented dialogues, emotional speech, and cross-lingual interactions to further broaden its applicability.

\section{Generative AI Use Disclosure}

During the preparation of this manuscript, the authors utilized generative AI tools to assist with manuscript editing, narrative polishing, and \LaTeX{} formatting troubleshooting. These tools were used strictly for structural and linguistic refinement. The AI tools were not used to generate the dataset, execute the experiments, or synthesize the core scientific claims. All authors have thoroughly reviewed the manuscript, consent to its submission, and assume full responsibility and accountability for the integrity and content of this work.

\section{Acknowledgements}
\label{sec:acknowledgements}

The project was funded through a grant from the Gates Foundation and a gift from Meta.

We thank all collaborators, institutions, and partners who contributed to the African Next Voices project. In particular, we acknowledge the support and engagement of Masakhane, Lelapa AI (\textit{Disclosure:} Vukosi Marivate is a co-founder of Lelapa AI), Lanfrica, the University of Pretoria (ITS and Research Office), Makerere University, and Digital Umuganda. We thank all of the collaborators on the larger project in Kenya (Maseno University and collaborators), Nigeria (Data Science Nigeria and collaborators), Mali (RobotsMali and collaborators) and beyond.

We also acknowledge the support of the ABSA Chair of Data Science and the Data Science for Social Impact (DSFSI) Lab at the University of Pretoria. Some authors were supported through the UK International Development and the International Development Research Centre (IDRC), Ottawa, Canada, under the AI4D Africa Programme. DSFSI further acknowledges gifts from NVIDIA, Google.org and OpenAI.

We are grateful to the Data Science for Social Impact Lab at UP, the professional staff at the University of Pretoria for their assistance. We also thank SADiLaR, the Agricultural Research Council, Grain SA, the Government of South Africa, Wikipedia, the African Wordnet Project, and others who made their data available for use in this project.

\bibliographystyle{IEEEbib}
\bibliography{refs}

\end{document}